\documentclass[runningheads]{llncs}

\usepackage{eccv}

\usepackage{eccvabbrv}

\usepackage{graphicx}
\usepackage{booktabs}

\usepackage{caption} %

\usepackage[accsupp]{axessibility}  %
\usepackage{algorithm}
\usepackage{algpseudocode}

\usepackage{hyperref} %

\usepackage{orcidlink}

\usepackage{array}
\usepackage{multirow}
\usepackage{multicol}
\usepackage{soul}

\newcommand{\shortname}{SMooDi\xspace}
\newcommand{\mypar}[1]{\vspace{1mm}\noindent\textbf{#1}}

\newif\ifdrafting
\draftingtrue %
\ifdrafting
    \newcommand{\yx}[1]{\textcolor{orange}{YX: #1}}
    \newcommand{\zl}[1]{\textcolor{cyan}{ZL: #1}}
    \newcommand{\hj}[1]{\textcolor{red}{HJ: #1}} 
    \newcommand{\ds}[1]{{\color{blue}[DS: #1]}}
    \newcommand{\VJ}[1]{{\color{magenta}[VJ: #1]}}

\else

	\newcommand{\yx} [1] {}
        \newcommand{\zl} [1] {}
	\newcommand{\hj} [1] {}
	\newcommand{\ds} [1] {}
        \newcommand{\VJ} [1] {}
 
\fi

\usepackage{amsmath,amsfonts,bm}

\def\eqref#1{equation~\ref{#1}}

\def\1{\bm{1}}

\def\vc{{\bm{c}}}

\def\vr{{\bm{r}}}
\def\vs{{\bm{s}}}

\def\vx{{\bm{x}}}

\def\vz{{\bm{z}}}

\DeclareMathAlphabet{\mathsfit}{\encodingdefault}{\sfdefault}{m}{sl}
\SetMathAlphabet{\mathsfit}{bold}{\encodingdefault}{\sfdefault}{bx}{n}

\newcommand{\vepsilon}{\mathbf{\epsilon}}

\begin{document}

\title{\shortname: Stylized Motion Diffusion Model}

\author{Lei Zhong\inst{1} \and
Yiming Xie\inst{1} \and
Varun Jampani \inst{2} \and 
Deqing Sun \inst{3} \and
Huaizu Jiang \inst{1}}

\authorrunning{Zhong et al.}

\institute{Northeastern University \and
Stability AI \and Google Research \\
\email{\{le.zhong,xie.yim,h.jiang\}@northeastern.edu} \\
\email{varunjampani@gmail.com} \\
\email{deqingsun@google.com}
}
\maketitle

\begin{abstract}

We introduce a novel Stylized Motion Diffusion model, dubbed \shortname, to generate stylized motion driven by content texts and style motion sequences.
Unlike existing methods that either generate motion of various content or transfer style from one sequence to another, \shortname can rapidly generate motion across a broad range of content and diverse styles.
To this end, we tailor a pre-trained text-to-motion model for stylization.
Specifically, we propose style guidance to ensure that the generated motion closely matches the reference style, alongside a lightweight style adaptor that directs the motion towards the desired style while ensuring realism.
Experiments across various applications demonstrate that our proposed framework outperforms existing methods in stylized motion generation. 
Project Page: \href{https://neu-vi.github.io/SMooDi/}{https://neu-vi.github.io/SMooDi/}


\end{abstract}

\section{Introduction}

\begin{figure}[t]
    \centering
       \includegraphics[width=0.95\linewidth]{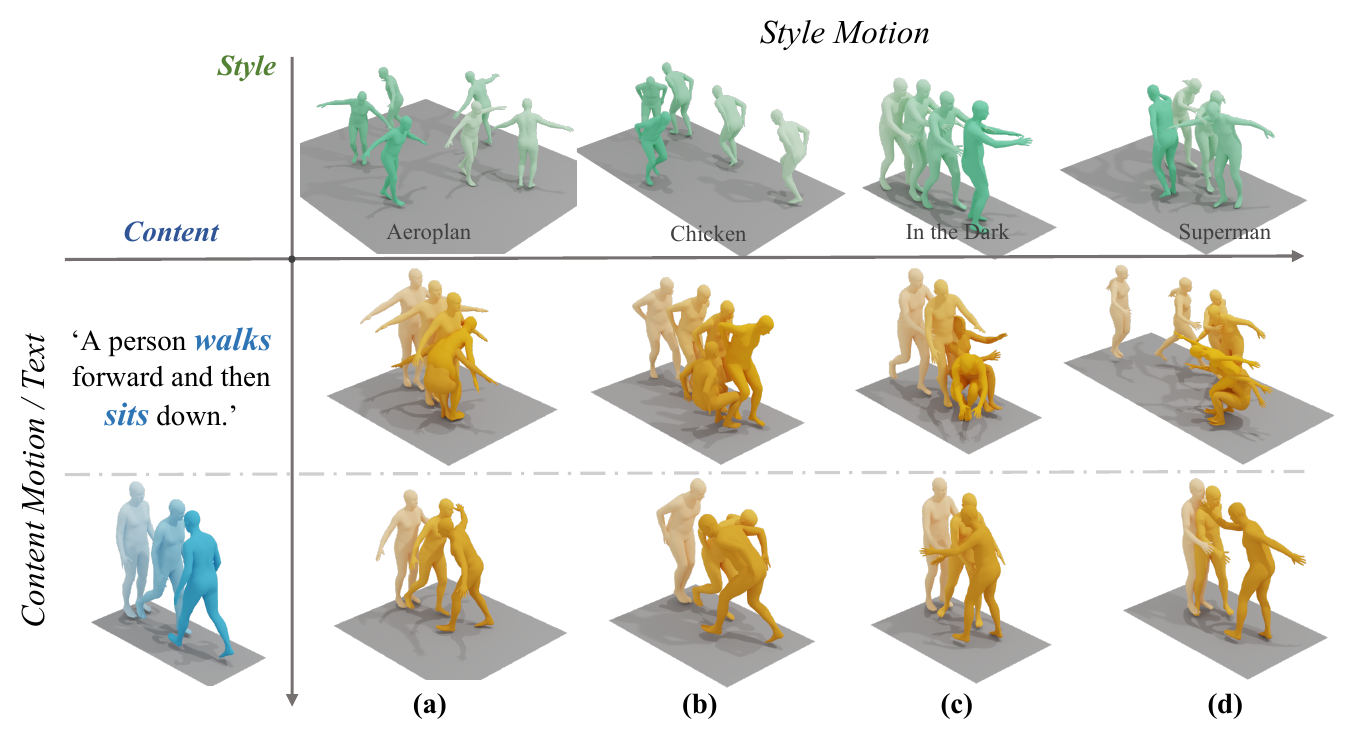}
       \caption{
            \textbf{\shortname can generate realistic, stylized human motions given a content text a style motion sequence.} It also accepts a motion sequence as content input. Darker color indicates later frames in the sequence. To better showcase the stylized motion generation, we place the style label for the each of the style motion sequence. Note that such style labels are not used as model input and shown here for visualization purpose only. 
            \textit{(Best viewed in color.)}
           }
       \label{fig:teaser}
\end{figure}

We address the problem of generating stylized motion from a content text and a style motion sequence, as shown in Fig.~\ref{fig:teaser}.
Human motion can typically be characterized  by two components: content and style.
Motion content represents the nature of a movement, such as walking and waving, and motion style reflects individual characteristics, such as personality traits (\eg, old, childlike) and emotions (\eg, happy, angry).
Traditional pipelines create stylized motions via motion capture from actors, and are both labor-intensive and time-consuming. 
Therefore, decades of research have focused on developing automatic methods to assist stylized motion creation~\cite{aberman2020unpaired,du2019stylistic,jang2022motion}.

Motion style transfer~\cite{aberman2020unpaired,tao2022style} is a practical and popular approach for the creation of stylized motion.
It transfers the style from an existing style motion sequence to another existing content motion sequence.
However, when a broad array of motion needs to be stylized, 
the pipeline may be inefficient -- it would first require the collection of a large number of  content motion sequences and then apply a motion style transfer method to process each sequence independently.
Moreover, motion sequences are not always readily available, especially for some customized content, such as running along a specific trajectory.
They may still need to be created first by actors or animators for stylization.

Recent advances of human motion generation with diffusion models~\cite{dhariwal2021diffusion,ho2022classifier} have shown impressive results of creating diverse and realistic human motions. 
But most efforts have concentrated on efficiently and accurately translating textual prompts into human motions, focusing on the \emph{content} only~\cite{petrovich2022temos,tevet2023human,chen2023executing}.
Integrating the \emph{style} condition to generate stylized motions remains under-explored.

Combining these two lines of research is a straightforward approach to tackle stylized motion generation, where a motion style transfer method~\cite{aberman2020unpaired, tao2022style} can be applied to each motion sequence generated by a text-driven motion diffusion model~\cite{petrovich2022temos,tevet2023human, chen2023executing}.
However, in addition to the aforementioned inefficiency issue, this approach has two more limitations.
First, error may accumulate across the pipeline. 
As, motion style transfer methods  are usually trained with high-quality real-world motion sequences,
we empirically observe that their performance may significantly degrade for imperfect motions produced by text-to-motion techniques. %
Second, existing motion style transfer methods rely on specialized style datasets~\cite{aberman2020unpaired,xia2015realtime,mason2022local} with limited motion content, which  
restricts their applications to motion diffusion models.

In this paper, we present a novel stylized motion diffusion model, dubbed \shortname{}, that customizes a pre-trained text-to-motion model for stylization.
Built upon the pre-trained motion latent diffusion model (MLD)~\cite{chen2023executing}, \shortname{} inherits MLD's ability to generate diverse motion content.
At the same time, \shortname{} can generate motions in a variety of styles according to different style reference conditions, as shown in Fig.~\ref{fig:teaser}.
Our main novelty is the style modulation module, which consists of a style adaptor and a style  guidance module. 
First of all, drawing inspiration from controllable image generation~\cite{zhang2023adding}, the style adaptor is designed to predict residual features conditioned on style reference motion sequence within each attention layer of MLD. 
It is useful for incorporating the style condition while ensuring the realism of the generated motion.
Second, we design both classifier-free and classifier-based style guidance to more precisely control the stylized motion generation.
Specifically, the classifier-free style and content guidance are linearly combined, where we can easily strike a balance between preserving content and reflecting style within the generated motion.
At the same time, we design an additional classifier-based style guidance mechanism.
It is an analytic function quantifying the disparity between the generated motion and the style reference motion in a style-centric embedding space, whose gradients are subsequently employed to guide the generated motion closer to the intended style.
Our style adaptor and guidance module are designed to be complementary, which lead to high-quality stylized motion generation.
The two modules are jointly optimized in a feature space instead of sequence-wise separate stitching, thereby avoiding the error accumulation issue.

Although our approach is primarily designed for stylized motion generation driven by content text, we can utilize DDIM-Inversion~\cite{song2020denoising} to identify the noisy latent corresponding to the content motion sequence. Following the same procedure as for text-driven content, \shortname is capable of facilitating stylized motion generation based on content motion sequences.
In other words, motion style transfer is a downstream application of our approach should it be desired in practice, \eg, to stylize the already created motion sequences.

Experiments on the HumanML3D~\cite{Guo_2022_CVPR} and 100STYLE~\cite{mason2022local} datasets demonstrate that \shortname surpasses other baseline models in generating stylized motion driven by content text, excelling in both content preservation and style reflection.
More importantly, unlike previous methods that require individual fine-tuning for each style~\cite{xu2023adaptnet, peng2021amp, everaert2023diffusion}, \shortname successfully integrates diverse content from the HumanML3D dataset and various styles from the 100STYLE dataset into a single model without requiring additional tuning during inference.

To summarize, our contributions are: (1) To our knowledge, \shortname is the first approach that adapts a pre-trained text-to-motion model to generate diverse stylized motion.
(2) We introduce a novel style modulation module that utilizes a stylized adaptor and a style classifier guidance to enable stylized motion generation while ensuring style reflection, content preservation, and realism.
(3)  Experiments demonstrate that \shortname not only sets a new state of the art in stylized motion generation driven by content text but also achieves performance comparable to state-of-the-art methods in motion style transfer.

\section{Related Work}

\subsection{Human Motion Generation}
Human motion generation has attracted great attention~\cite{Guo_2022_CVPR,guo2024momask,pinyoanuntapong2024mmm,pinyoanuntapong2024bamm,wan2023tlcontrol,wu2024thor,petrovich2024multi,wu2024motionllm,chen2024motionllm,dai2024motionlcm,raab2024monkey,xu2024interdreamer,cen2024generating,yi2024generating,cohan2024flexible,jiang2023motiongpt,zhang2023motiongpt}.
Inspired by the impressive performance of diffusion models in image generation, a lot of works~\cite{tevet2023human,chen2023executing,zhang2024motiondiffuse,yuan2023physdiff,karunratanakul2023gmd,rempeluo2023tracepace,huang2023diffusion,kulkarni2023nifty,xu2023interdiff,Pi_2023_ICCV,dabral2023mofusion,shafir2023human,peng2023hoidiff,xie2023omnicontrol,zhou2023emdm,petrovich2024multi,wang2023intercontrol,karunratanakul2023dno,ghosh2023remos} utilize diffusion models to generate human motion.
MDM~\cite{tevet2023human} facilitates high-quality generation and versatile conditioning, providing a solid baseline for novel motion generation tasks.
MLD~\cite{chen2023executing} minimizes computational overhead during both training and inference by establishing the diffusion process within the latent space.
Driven by the efficacy of diffusion models for control and conditioning, several studies have leveraged pre-trained motion diffusion models to generate long-sequence motions~\cite{shafir2023human}, enable human-object interactions~\cite{peng2023hoidiff}, and control the joint trajectory of generated motions~\cite{karunratanakul2023gmd, xie2023omnicontrol}.
However, there is no work exploring how to leverage pre-trained motion diffusion models to generate diverse stylized motion.
While some studies~\cite{alexanderson2023listen,Ao2023GestureDiffuCLIP,raab2023single} have enabled stylized motion generation in their diffusion pipeline, their methods are trained from scratch, and the supported styles are restricted by their motion content dataset.
It is challenging for them to simultaneously support diverse motion content and style.
In this work, we build upon a pre-trained motion diffusion model, MLD, and explore how to fine-tune it on a larger motion style dataset, 100STYLE, to learn diverse motion styles while retaining the ability to support motion generation across a wide range of content.

\subsection{Motion Style Transfer}
Recently, motion style transfer has seen quality enhancements through the adoption of various advanced neural architectures and generative models, such as graph neural networks~\cite{park2021diverse}, time-series models~\cite{tao2022style,mason2022local}, normalizing flows~\cite{wen2021autoregressive}, and diffusion models~\cite{Ao2023GestureDiffuCLIP,raab2023single}.
Specifically, Aberman et al.~\cite{aberman2020unpaired} designed a two-branch generative adversarial network to disentangle motion style from content and facilitate their re-composition.
Their approach effectively breaks the constraint of requiring a paired motion dataset.
Motion Puzzle~\cite{jang2022motion} realizes a framework that can control the style of individual body parts.
Above methods extract both content and style features from the motion sequence.
Moreover, Guo et al.~\cite{guo2024generative} leverage the latent space of pre-trained motion models to enhance the extraction and infusion of motion content and style. %
However, a major limitation of these models is their reliance on specialized style datasets~\cite{aberman2020unpaired,xia2015realtime} with limited motion content, which restricts their applications.
In this work, we customize a pre-trained text-to-motion model for stylization, thus inheriting its ability to generate diverse motion content.

\begin{figure}[t]
    \centering
       \includegraphics[width=0.9\linewidth]{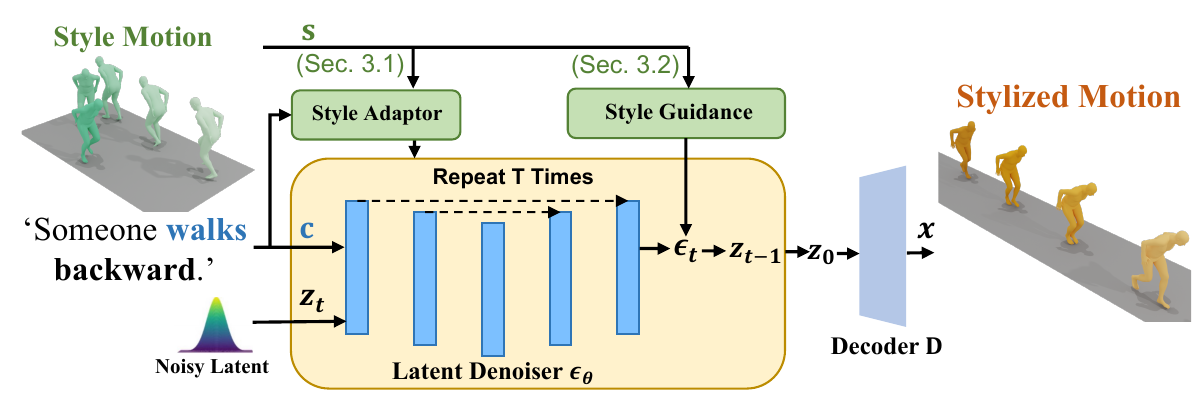}
       \caption{
            \textbf{Overview of \shortname}.
            Our model generates stylized human motions from content text and a style motion sequence. At the denoising step $t$, our model takes the content text $\mathbf{c}$, style motion $\mathbf{s}$, and noisy latent $\mathbf{z}_t$ as input and predicts $\epsilon_t$, which is then transferred to $\vz_{t-1}$. This denoising step is repeated $T$ times to obtain the noise-free motion latent $\mathbf{z}_0$, which is fed into a motion decoder $D$ to produce the  stylized motion. 
           }
       \label{fig:overview}
\end{figure}
\section{Stylized Motion Diffusion Model}
In this section, we introduce our proposed \shortname for incorporating style conditions from a style motion sequence into a content-oriented pre-trained motion diffusion model (MLD~\cite{chen2023executing}).
Fig.~\ref{fig:overview} presents an overview of \shortname. 
Following the setting in MLD~\cite{chen2023executing}, we place the diffusion process in the motion latent space. 
Let $\vepsilon_\theta$ denote the latent denoiser (a UNet parameterized by $\theta$), and $\{\vz_t\}_{t=0}^T$ denote the sequence of noisy latents, where $\vz_T$ is a Gaussian noise.
Given a content prompt $\mathbf{c}$ and a style prompt $\mathbf{s}$, we define $\epsilon_{t} = \epsilon_\theta (\vz_t, t, \mathbf{c}, \mathbf{s})$ for the denoising at step $t~(0 < t \leq T)$. A cleaner noisy latent $\vz_{t-1}$ can be obtained by subtracting $\epsilon_{t}$ from $\vz_t$.
The denoising step is repeated $T$ iterations until a clean latent $\vz_0$ is obtained.
It can then be decoded by a motion decoder $\mathbf{D}$ into a realistic motion sequence $\vx \in \mathbb{R}^{N \times H}$ that accurately reflects both the content and style conditions.
Here, $N$ represents the length of the motion sequence, and $H$ is the dimension of human motion representations. We employ the same motion representations as in HumanML3D~\cite{Guo_2022_CVPR}, where $H=263$.

As shown in the Fig.~\ref{fig:overview}, the content prompt is a text description, and the style prompt is provided by a reference style motion sequence $\mathbf{s} \in \mathbb{R}^{N \times H}$.
In this section, we focus on using a text description as the content prompt $\mathbf{c}$ to explain our proposed stylized motion diffusion model.
By employing the DDIM-Inversion~\cite{song2020denoising} to identify the noisy latent corresponding to a motion content sequence, we can effectively use motion sequences as content prompts to generate stylized motions.
In other words, motion style transfer is a downstream application of our proposed approach. 

Our proposed stylization module consists of a style adaptor and a style guidance module. We will explain them separately in the rest of this section.

\subsection{Style Adaptor}
\label{sec:approach}

\begin{figure}[t]
    \centering
       \includegraphics[width=0.98\linewidth]{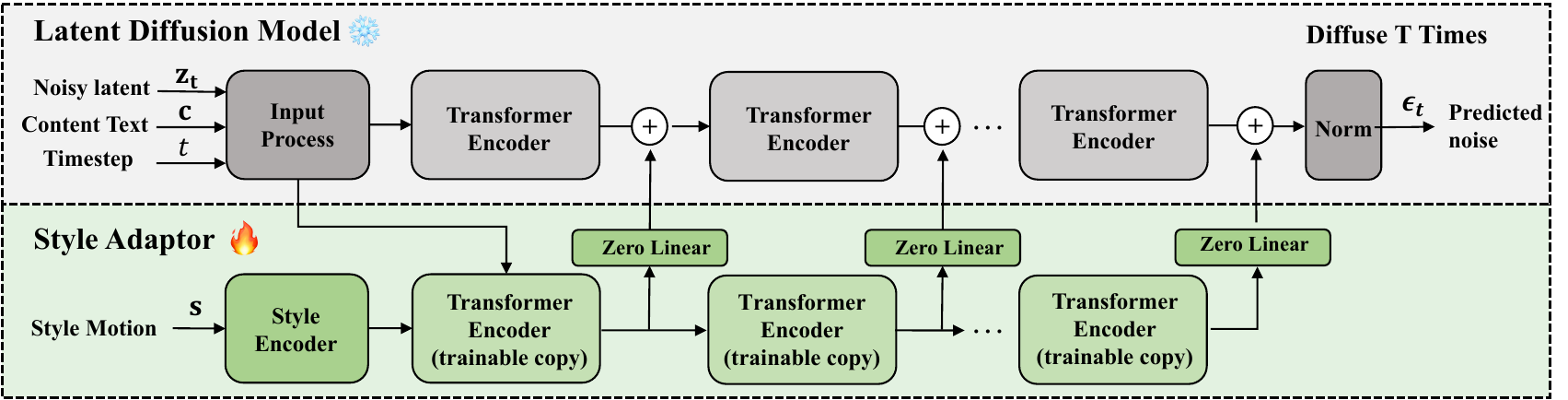}
       \caption{
            \textbf{Detailed illustration of our proposed style adaptor}. The style adaptor is connected to the motion diffusion model via zero linear layer. The output of the style adaptor from each Transformer encoder is added to the motion diffusion model to steer the predicted noise towards the target style.
           }
       \label{fig:pipeline}
\end{figure}
Although LoRA has been successfully used to incorporate ``style'' into the models in image domain~\cite{shah2023ziplora,jones2024customizing}, they typically require training a separate LoRA for each style. 
In contrast, we focus on fine-tuning the model just once to adapt to various motion styles, where adapting ControlNet~\cite{zhang2023adding} is more suitable.
Therefore, we design a content-aware style adaptor based on ControlNet. This adaptor incorporates the motion style condition into the pre-trained MLD~\cite{chen2023executing}. 

Instead of learning to disentangle motion style from content from scratch on a large motion dataset, we redirect our focus towards capturing the motion style while ensuring the preservation of diverse motion content within the pre-trained MLD framework.
Specifically, it consists of a trainable copy of the Transformer Encoder from the latent diffusion model in MLD. The architecture of the style adaptor is illustrated in Fig.~\ref{fig:pipeline}. 
An independent style encoder is utilized to extract the style embedding from the style motion sequence $\mathbf{s}$. 
The style adaptor takes the same content prompt $\mathbf{c}$, the noised latent $\vz_t$ and timestep $t$ as in MLD, and the extracted style embedding. 
Each Transformer layer in the original latent diffusion model and the style adaptor is connected via a linear layer, with both weight and bias initially set to zeros. As training progresses, the style adaptor learns the style constraints and gradually applies the learned feature corrections to the corresponding layers in the latent diffusion model, thereby implicitly steering the output towards the desired style.

\subsection{Style Guidance}
\label{section:sec}
The style adaptor alone may not be sufficient to successfully incorporate the style condition.
We further leverage both the classifier-free and classifier guidance to further enhance the stylization of a motion diffusion model.
The combination of two types of guidance effectively ensures that generated motion meets multiple constraints while maintaining realism, complementing the style adaptor.

\noindent \textbf{Classifier-free Style Guidance.}
With the introduction of an extra style condition, we can divide the conditioned classifier-free guidance into two parts.
\begin{align}
\label{eql:cfg}
 \mathbf{\epsilon}_\theta(\vz_t, t, &\mathbf{c},\mathbf{s})  =  \mathbf{\epsilon}_\theta(\vz_t, t, \emptyset, \emptyset) + \notag\\ 
 & \underbrace{w_c (\mathbf{\epsilon}_\theta(\vz_t, t, \mathbf{c}, \emptyset) -  \mathbf{\epsilon}_\theta(\vz_t, t, \emptyset, \emptyset) )}_{\text{Classifier-free Content Guidance}} +  \underbrace{w_s (\mathbf{\epsilon}_\theta(\vz_t, t,\mathbf{c}, \mathbf{s}) -  \mathbf{\epsilon}_\theta(\vz_t, t,\mathbf{c}, \emptyset) )}_{\text{Classifier-free Style Guidance}},
\end{align}
where $w_c$ and $w_s$ represent the strengths of the classifier-free guidance for the condition $\mathbf{c}$ and $\mathbf{s}$, respectively.
We slightly abuse the notations here by using $\emptyset$ to denote a condition is not used.
The classifier-free content guidance is the same as in MLD, which can facilitate the text-to-motion generation process in combination with the first term's unconditioned guidance.
Our proposed classifier-free style guidance works in a similar way.
Note that $\vepsilon_{\theta}(\vz_t, t, \vc, \vs)$ is MLD model with the style adaptor incorporated introduced in the previous section, which takes both a textual prompt and style motion sequence as input.
By contrasting the text-driven denoising output with and without the style condition, it can highlight the effectiveness of the style input $\vs$ and facilitate the generation of stylized motion driven by content text.
Our insight here is that by dividing the conditioned guidance into content and style components separately, we can easily strike a balance between preserving content and reflecting style within the generated motion.

\begin{figure}[t]
    \centering
       \includegraphics[width=0.9\linewidth]{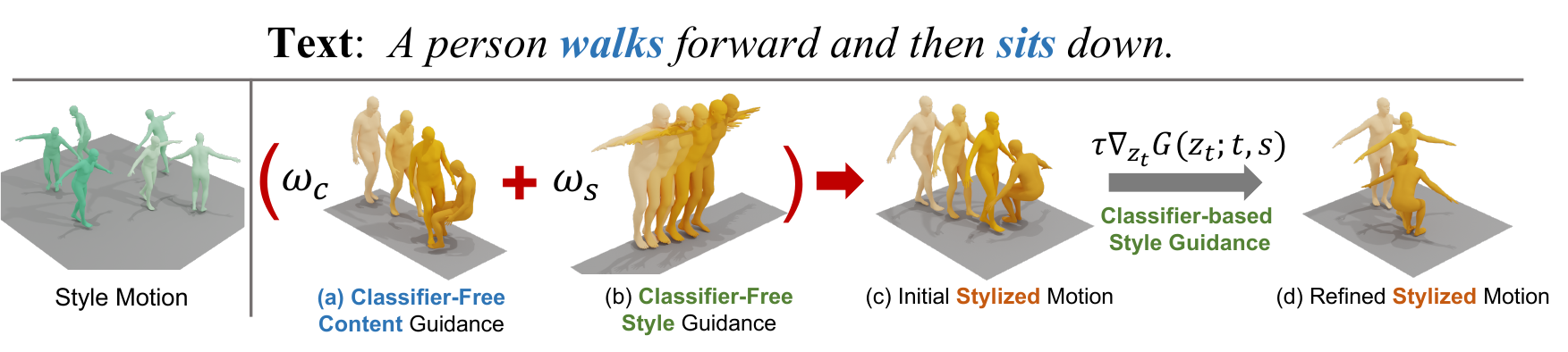}
       \caption{
            \textbf{Visual illustrations of the classifier-free and clasifier-based style guidance.} (a) and (b) respectively show the classifier-free content and style guidance;
            (c) displays the initial stylized motion resulting from the combination of (a) and (b); (d) illustrates the refined stylized motion modified by the classifier-based style guidance.
           }
       \label{fig:visual_guidance}
\end{figure}

To better understand the classifier-free content and style guidance, we visualize each of them through decoding denoised latent $\vz_0$ into the motion space. 
As illustrated in Fig.\ref{fig:visual_guidance}(a), the content guidance ensures the motion generation is faithful to the textual prompt, while the style  guidance, as shown in Fig.\ref{fig:visual_guidance}(b), emphasizes style-related characteristics in the output.
Combining both forms of guidance results in a stylized motion that adheres to both content and style conditions, as illustrated in Fig. \ref{fig:visual_guidance}(c).

\noindent \textbf{Classifier-based Style Guidance.}
To further improve the stylization of a motion diffusion model, we adopt the classifier guidance~\cite{dhariwal2021diffusion,yu2023freedom} to provide stronger guidance to the generated motion towards the desired style. 
The core of our classifier-based style guidance is a novel analytic function $G(\vz_t, t, \mathbf{s})$, which calculates the $L_1$ distance between the style embedding of the generated clean motion $\hat{\vx}_0$ at denoising step $t$ and the reference style motions $\mathbf{s}$. 
The function's gradient is utilized to steer the generated motion towards the desired style.
\begin{align}
\mathbf{\epsilon}_\theta(\vz_t, t, \mathbf{c}, \mathbf{s}) &= \mathbf{\epsilon}_\theta(\vz_t, t, \mathbf{c},\mathbf{s}) + \tau \nabla_{\vz_t}G(\vz_t, t, \mathbf{s}), \notag\\
G(\vz_t, t, \mathbf{s}) &= |f(\hat{\vx}_0) - f(\mathbf{s})|,
\end{align}
where $\tau$ adjusts the strength of reference-style guidance and $f$ denotes the style feature extractor.
The generated motion $\hat{\vx}_0$ is obtained by first converting the denoising output latent $\vz_t$ into the predicted clean latent as shown below:
\begin{equation}
\hat{\vz}_0 =  \frac{\vz_t - \sqrt{1 - \alpha_t} \varepsilon_\theta(\vz_t, t, \vc, \vs)}{\sqrt{\alpha_t}},
\end{equation}
where $\alpha_t$ denotes the pre-defined noise scale in the forward process of the diffusion model. The predicted clean latent $\hat{\vz}_0$ is then input into the motion decoder $\mathbf{D}$ to obtain the generated motion.

We obtain the style feature extractor by training a style classifier on the 100STYLE dataset~\cite{mason2022local} and removing its final layer. We refer the readers to more details in the supplementary materials.
The training of the style feature extractor with ground-truth style labels for supervision enables it to effectively capture style-related features.
Therefore, style classifier guidance can provide more guidance to the stylized motion generation.

\noindent \textbf{Combination of the Two Style Guidance.}
The classifier-free and classifier-based style guidance are designed to complement each other, each playing a vital role in accurately reflecting the target style in the generated motions. 
As illustrated in Fig.~\ref{fig:visual_guidance}, the desired style motion is ``arms open wide to the sides like an airplane''. 
The classifier-free style guidance (Fig.~\ref{fig:visual_guidance}(b)) can capture style-related characteristic in a reasonably accurate manner.
When combined with the classifier-free content guidance, it depicts the desired style (Fig.\ref{fig:visual_guidance}(c)). 
Refined further by the classifier-based style guidance, the stylization is more authentic, where the person's harms are more open (Fig.\ref{fig:visual_guidance}(d)).
In addition to such visual results, quantitative ablation studies also verify the effectiveness of our proposed both classifer-free and classifer-based style guidance.

At the same time, although classifier-based style guidance offers precise style control, its effectiveness may be compromised when the content text significantly diverges from locomotion-related movements. This is because the style feature function is trained solely on the 100STYLE dataset, which contains only such movements. 
Over-reliance on style classifier guidance risks producing motions that fail to execute the desired actions, leading to unrealistic and physically implausible movements.
Therefore, we leverage a content-aware style adaptor that establishes the fundamental style direction, while style classifier guidance refines this base for a more precise outcome. 
The effectiveness of this design is verfied in our ablation studies.

\subsection{Learning Scheme}
\label{section:loss}
Following~\cite{zhang2023adding}, a straightforward approach to train \shortname is to freeze the parameters of MLD and solely train the style adaptor on the 100STYLE dataset using the following loss function:
\begin{equation}
\mathcal{L}_{std} = \mathbb{E}_{\epsilon, \vz} \left[ \left\| \epsilon_{\theta}(\vz_t, t, \mathbf{c}, \mathbf{s}) - \epsilon \right\|^2_2 \right],
\end{equation}
where $\epsilon \sim \mathcal{N}(0, \mathbf{I})$ represents the ground-truth noise added to $\vz_0$ .
In our experiments, however, we found that this loss function alone leads to an issue of ``content-forgetting'', where the model progressively looses the MLD's ability to generate motions with diverse contents.
To address this issue, we design a content prior preservation loss $L_{pr}$.
Specifically, we randomly sample motions from the HumanML3D dataset to compute a prior preservation loss when fine-tuning \shortname on the 100STYLE dataset.
\begin{equation}
\mathcal{L}_{pr} = \mathbb{E}_{\epsilon^{\prime},\vz^{\prime}} \left[ \left\| \epsilon_{\theta}(\vz_t^{\prime}, t, \mathbf{c}^{\prime}, \mathbf{s}^{\prime}) - \epsilon^{\prime} \right\|^2_2 \right],
\end{equation}
where $\vz_t^{\prime}$, $\mathbf{c}^{\prime}$ and $\mathbf{s}^{\prime}$ represents the motion latent, content prompt and style motion sequence derived from the HumanML3D dataset.
$\epsilon^{\prime}$ is the noise map added to $\vz^{\prime}_0$.
A similar solution is used in DreamBooth~\cite{ruiz2023dreambooth} to solve the ``language drift'' problem, where images generated from the frozen pretrained image generation model are utilized to enforce a class-prior preservation loss during model fine-tuning.
Our content preservation loss can effectively mitigate content forgetting while learning diverse motion styles from the 100STYLE dataset.

To further encourage the style adaptor to focus on motion style, while also ensuring that the latent diffusion model in MLD handles motion content well, we introduce an additional cycle prior-preservation loss, inspired by~\cite{jang2022motion,xu2024cyclenet}.
Specifically, we start this process by randomly sampling content text and motion style sequences from both the 100STYLE and HumanML3D datasets simultaneously. Then, we intermix the content text and motion style from these sequences with each other. Finally, we repeat this process to reconstruct the original motion sequences.
The formula is expressed as follows:
\begin{equation}
\mathcal{L}_{cyc} = \mathbb{E}_{\vz,\vz^{\prime}, \epsilon, \epsilon^{\prime}} \left[ \left\| \epsilon_{\theta}(\vz_t^{sh}, t, \mathbf{c}, \vs^
{hs}) + \epsilon_{\theta}(\vz_t^{hs}, t, \mathbf{c}^{\prime}, \vs^{sh}) - \epsilon - \epsilon^{\prime} \right\|^2_2 \right],
\end{equation}
where $\vs^{hs}$ denotes the motion sequence created by merging content from the HumanML3D dataset with style from the 100STYLE dataset.
Similarly, $\vs^{sh}$ represents the sequence where content is sourced from the 100STYLE dataset and style from the HumanML3D dataset.
The noised latent codes $\vz_{t}^{sh}$ and $\vz_{t}^{hs}$ correspond to $\vs^{sh}$ and $\vs^{hs}$, respectively.
Essentially, the cycle prior-preservation loss exchanges diverse content and style between two datasets, encouraging the content text to remain invariant in the generated motion under forward and backward translation.
Overall, the training loss function of our framework is defined as follows:
\begin{equation}
\mathcal{L}_{all} = \mathcal{L}_{std} + \lambda_{pr} \mathcal{L}_{pr} + \lambda_{cyc} \mathcal{L}_{cyc}
\end{equation}
where $\lambda_{pr}$ and $\lambda_{cyc}$ are hyperparameters. 
We refer readers to the pseudocode and illustration for training in the supplementary material for more details.

\begin{table*}[t]
\centering
\caption{Comparison with baseline methods on stylized motion generation driven by content text, using a combination of the 100STYLE (providing style) and HumanML3D datasets (providing content).}
\label{tab:compare_with_baseline}
\resizebox{0.85\textwidth}{!}{
\begin{tabular}{c|cccccccc}
\toprule
Method  & {FID $\downarrow$} & \multicolumn{1}{p{1.8cm}}{\centering Foot skating \\ ratio $\downarrow$} & {MM Dist $\downarrow$}  &\multicolumn{1}{p{1.9cm}}{\centering R-precision $\uparrow$ \\ (Top-3)}& {Diversity $\rightarrow$}& {SRA $\uparrow$}  \\ 
\midrule
Ours &  1.609 & \textbf{0.124} & 4.477 & 0.571 & \textbf{9.235} & \textbf{72.418} \\
\midrule
MLD+Motion Puzzle~\cite{jang2022motion} & 6.127 & 0.185 & 6.467 &0.290& 6.4762 & 63.769  \\
MLD+Aberman et al.~\cite{aberman2020unpaired} & 3.309 & 0.347 & 5.983 & 0.406 & 8.816 & 54.367 \\
ChatGPT+MLD& \textbf{0.614} & 0.131 & \textbf{4.313} & \textbf{0.605} & 8.836 & 4.819\\
\bottomrule
\end{tabular}
}
\end{table*}
\begin{figure}[t]
    \centering
       \includegraphics[width=0.9\linewidth]{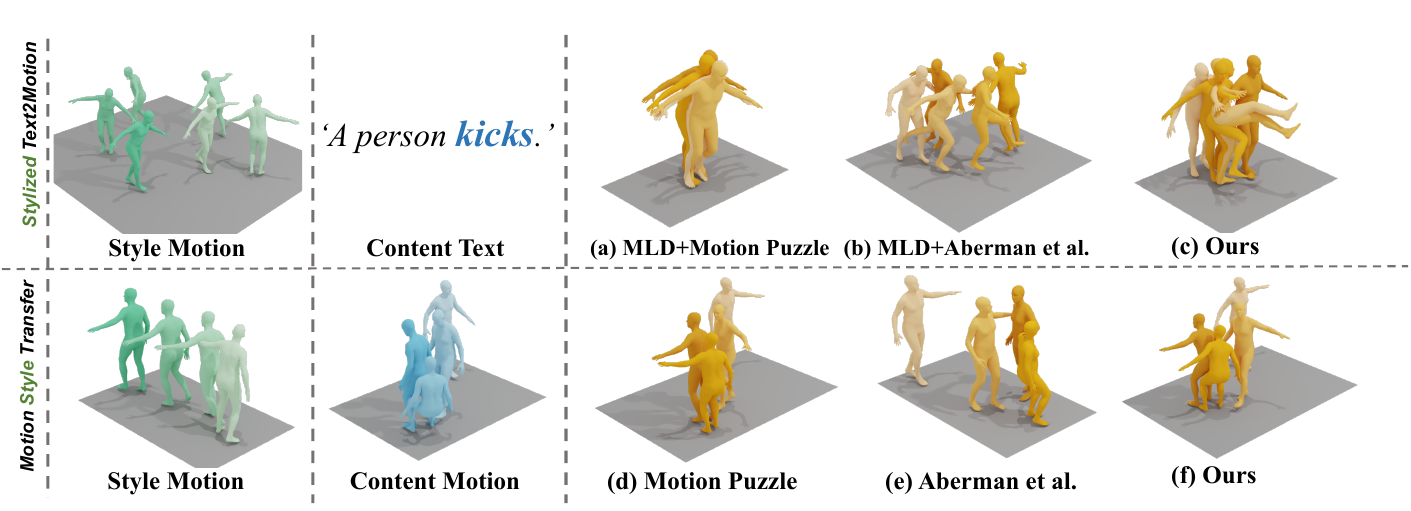}
       \caption{
           Qualitative comparisons of our approach and baseline methods on two stylized motion generation task.
           }
       \label{fig:qualitative}
\end{figure}
\section{Experiments}
We conduct experiments on both stylized text2motion and motion style transfer to demonstrate the effectiveness of our framework. 
Both tasks use a motion sequence as a style prompt, with the primary difference being their content prompt input: the former utilizes text, while the latter relies on motion sequences.

\noindent \textbf{Datasets.}
We utilize the HumanML3D dataset~\cite{Guo_2022_CVPR} as our motion content dataset and the 100STYLE dataset~\cite{mason2022local} as our motion style dataset. The HumanML3D dataset is the largest motion capture dataset, featuring text annotations and comprising 14,646 motions and 44,970 motion annotations.
Following the processing approach outlined in~\cite{Guo_2022_CVPR}, we preprocess the HumanML3D dataset to obtain consistent motion representations.
On the other hand, the 100STYLE dataset~\cite{mason2022local}, being the largest motion style dataset, comprises up to 1,125 minutes of motion sequences, showcasing a wide array of 100 diverse locomotion styles.
Due to differences in skeletons between the 100STYLE dataset and HumanML3D, we retarget the motions from 100STYLE to match the HumanML3D (SMPL-H) skeleton. Following this alignment, we apply the same processing steps as used for the HumanML3D dataset to preprocess the 100STYLE dataset.
Moreover, as the 100STYLE dataset lacks text descriptions, we leverage MotionGPT~\cite{jiang2023motiongpt} to generate pseudo text descriptions for the motion sequences in the 100STYLE dataset. 

\noindent \textbf{Evaluation metrics} are designed to assess three dimensions: Content Preservation, Style Reflection, and Realism.
For content preservation and style reflection assessment, we employ metrics consistent with those used in~\cite{chen2023executing}: motion-retrieval precision (R precision), Multi-modal Distance (MM Dist), Diversity, and Frechet Inception Distance (FID).
Additionally, recognizing the common foot skating issues in kinetics-based motion generation methods, we incorporate the foot skating ratio metric proposed by~\cite{karunratanakul2023gmd} into our motion quality evaluation.
For style reflection, we employ Style Recognition Accuracy (SRA)~\cite{jang2022motion}. During evaluation,  we randomly select a content  text from the HumanML3D dataset and a motion style sequence from the 100STYLE dataset to generate the stylized motion. We then use a pre-trained style classifier to compute the SRA for the generated motion.
It's noteworthy that some motion style labels in the 100STYLE dataset, like 'kick' and 'jump,' inherently convey motion content, which may conflict with the content text in HumanML3D dataset. To address this, we categorize the motion style labels into server groups following the approach by Kim et al.~\cite{kim2019perceptual}, Specifically excluding the 'ACT' group ensures that only motion style labels not conflicting with motion content are considered when computing the SRA metric.
Further details about the evaluation metrics are provided in the supplementary material.

\noindent \textbf{Baselines.} 
For motion style transfer task, we compare our methods with two state-of-the-art methods, namely Motion Puzzle~\cite{jang2022motion} and Aberman et al.~\cite{aberman2020unpaired}. To ensure a fair comparison, we train the compared methods under the same settings as ours, using a combined dataset comprising HumanML3D and 100STYLE.
Due to the constraints of the multi-class discriminator in Aberman et al., which requires style labels, we adopt the training method outlined in Motion Puzzle to eliminate the need for style labels.
For stylized text2motion task, we compare our method against baselines capable of generating stylized motion from content text and style motion sequences. The straightforward baselines involve applying motion style transfer methods to the motion sequences generated by the text2motion model.
To align with our approach that uses a pre-trained motion diffusion model, the text2motion models in the baselines for stylized motion generation select MLD, and the motion style transfer methods are consistent with those used in the motion style transfer task.
For the stylized text2motion task, we compare our method against two kinds of baselines capable of generating stylized motion from content text and style motion sequences. The first kind of baseline involves applying motion style transfer methods to the motion sequences generated by the text2motion model. To align with our approach that uses a pre-trained motion diffusion model, the text2motion models select MLD~\cite{chen2023executing}, and the motion style transfer methods are consistent with those used in the motion style transfer task.
The second kind of baseline involves using ChatGPT to merge style labels from 100STYLE with text from HumanML3D into a sentence. For example, given the content text 'a person walks.' and the style label 'old,' we obtain 'an elderly person walks.' This merged sentence is then fed to MLD.

\begin{table*}[t]
\centering
\caption{Comparison with baseline methods on motion style transfer.}
\label{tab:compare_sota_mst}

\begin{subtable}{.5\textwidth}
\caption{Evaluation on HumanML3D dataset}
\centering
\resizebox{!}{0.90cm}{
\begin{tabular}{@{}lccc@{}}
\toprule
Method & \multicolumn{1}{p{1.8cm}}{\centering Foot skating \\ ratio $\downarrow$} & FID $\downarrow$ & SRA$\uparrow$(\%) \\
\midrule
Ours & \textbf{0.095} & \textbf{1.582} & 65.147 \\
Motion Puzzle~\cite{jang2022motion} & 0.197 & 6.871 & \textbf{67.233} \\
(Aberman et al~\cite{aberman2020unpaired}) & 0.338 & 3.892 & 61.006 \\
\bottomrule
\end{tabular}
}
\end{subtable}%
\begin{subtable}{.5\textwidth}
\caption{Evaluation on Xia dataset}
\resizebox{!}{0.88cm}{
\centering
\begin{tabular}{@{}lcccc@{}}
\toprule
Method & \multicolumn{1}{p{1.8cm}}{\centering Foot skating \\ ratio$\downarrow$} & FID$\downarrow$ & SRA$\uparrow$(\%) & CRA$\uparrow$(\%) \\
\midrule
Ours & 0.0317 & \textbf{4.663} & 61.111 & \textbf{45.555} \\
Motion Puzzle~\cite{jang2022motion} & 0.0316 & 5.360 & \textbf{67.778} & 25.556 \\
(Aberman et al~\cite{aberman2020unpaired}) & \textbf{0.0260} & 5.681 & 56.667 & 34.444 \\
\bottomrule
\end{tabular}
}
\end{subtable}

\end{table*}

\subsection{Comparison to Baseline Methods}
\noindent \textbf{Quantitative and Qualitative}
For the task of stylized text2motion, Table~\ref{tab:compare_with_baseline} reports the comparisons of our method with the three baseline methods.

As shown in the $3^{rd}$ row of Table~\ref{tab:compare_with_baseline}, ChatGPT+MLD only achieves around \textbf{$5.29\%$} in terms of SRA, indicating that MLD cannot enable stylized generation from text alone, even though it contains style descriptions.
Notably, our method outperforms the two baselines that combine MLD with motion style transfer methods in all metrics.

Specifically, our method performs better than MLD+Motion Puzzle in the SRA metric by 13.56\% and significantly outperforms MLD+Aberman et al. in the FID metric by \textbf{51.38\%} and \textbf{0.64\%} in the R-precision metric.
The first row of Fig.~\ref{fig:qualitative} validates our observation, where the motion generated by our method performs better in adhering to both content and style constraints than baseline methods.
In contrast, MLP+Aberman et al. can successfully perform the action but fail to reflect the motion style in Fig.\ref{fig:qualitative}(b), while Motion Puzzle can accurately reflect the motion style but struggles to effectively perform the action in Fig.\ref{fig:qualitative}(a).

For the task of motion style transfer, since it does not take content text as input, text-motion related metrics such as MM Dist, R-precision, and Diversity are not applicable and thus are not reported.
Part (a) of Table~\ref{tab:compare_sota_mst} presents a comparison between our method and the two baseline methods, using the HumanML3D dataset as the motion content source and drawing motion styles from the 100STYLE dataset.
Our method delivers competitive results in the SRA metric and excels in the FID and foot skating ratio metrics. 
Specifically, we see a substantial \textbf{59.35\%} improvement in the FID metric over Aberman et al.~\cite{aberman2020unpaired}.
To more effectively compare the generalizability of different methods, we conduct experiments on the Xia dataset~\cite{xia2015realtime}, a small, specific motion style dataset that was unseen by our and the baseline models during training. Because motion content labels are present in the Xia dataset, we report the Content Recognition Accuracy (CRA). Part (b) of Table~\ref{tab:compare_sota_mst} showcases the results. Our method maintains competitive performance in the SRA metric, with only a marginal 9.83\% decrease in SRA compared to Motion Puzzle.
On the contrary, our method exhibits a significant 32.26\% increase in the CAR metric relative to Aberman et al., and a notable 78.26\% enhancement over Motion Puzzle. Our method achieves a better balance between style reflection and content preservation. The second row of Fig.~\ref{fig:qualitative} validates this observation. It is worth noting that, unlike other motion style transfer methods, our method does not incorporate objectives for enabling stylized motion generation using motion content sequences. Through simple DDIM-Inversion and without any additional optimization or regularization, our method achieves performance comparable to existing motion style transfer methods.

\begin{table*}[t]
\centering
\caption{Ablation Studies on HumanML3D Content and 100STYLE Styles.
}
\label{tab:ablation}
\resizebox{0.75\textwidth}{!}{
\begin{tabular}{c|cccccccc}
\toprule
Method  & {FID$\downarrow$} & \multicolumn{1}{p{1.8cm}}{\centering Foot skating \\ ratio$\downarrow$} & {MM Dist$\downarrow$}  &\multicolumn{1}{p{1.9cm}}{\centering R-precision$\uparrow$ \\ (Top-3)}& {Diversity$\rightarrow$}& {SRA(\%)$\uparrow$}  \\ 
\midrule
Ours (on all) & 1.609 & 0.124 & 4.477 & 0.571 & 9.235 & 72.418 \\
\midrule
$w/o \; L_{cyc}$  & 2.046 & 0.136 & 4.465 & 0.569 & 8.869 & 64.866 \\ 
\(w/o \, L_{pr}+L_{cyc}\) & 5.996 & 0.166 & 6.098 & 0.335 & 7.456 & 81.841 \\ 

\midrule
\textit{w/o classifier-based} & 1.050 & 0.111 & 4.085 & 0.630 & 9.445 & 20.245 \\
\textit{w/o adaptor}  & 2.984 & 0.123 & 4.526 & 0.550  & 8.372 & 69.952 \\

\bottomrule
\end{tabular}
}
\end{table*}

\begin{figure}[t]
    \centering
       \includegraphics[width=0.9\linewidth]{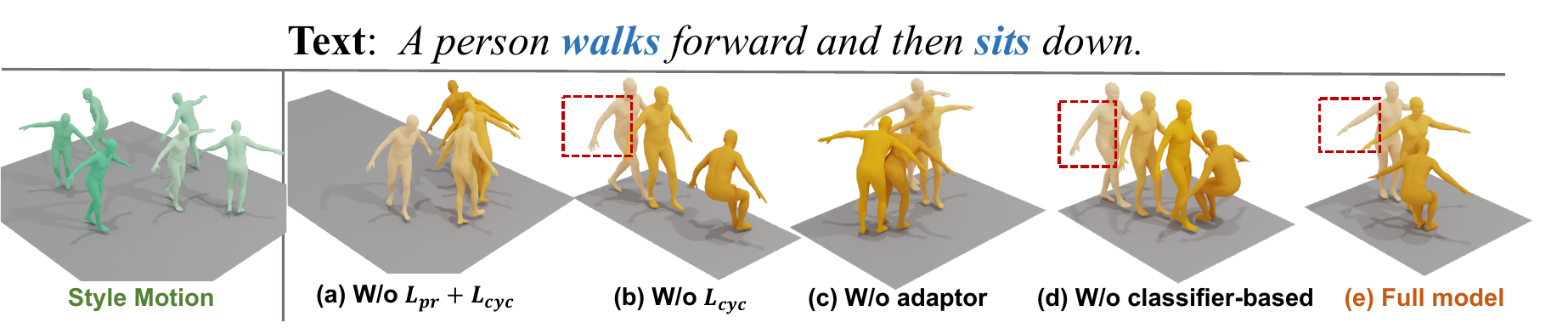}
       \caption{
            \textbf{Visual comparisons} of the ablation designs and our full model.
           }
       \label{fig:ablation}
\end{figure}
\begin{figure}[tp]
    \centering
       \includegraphics[width=0.8\linewidth]{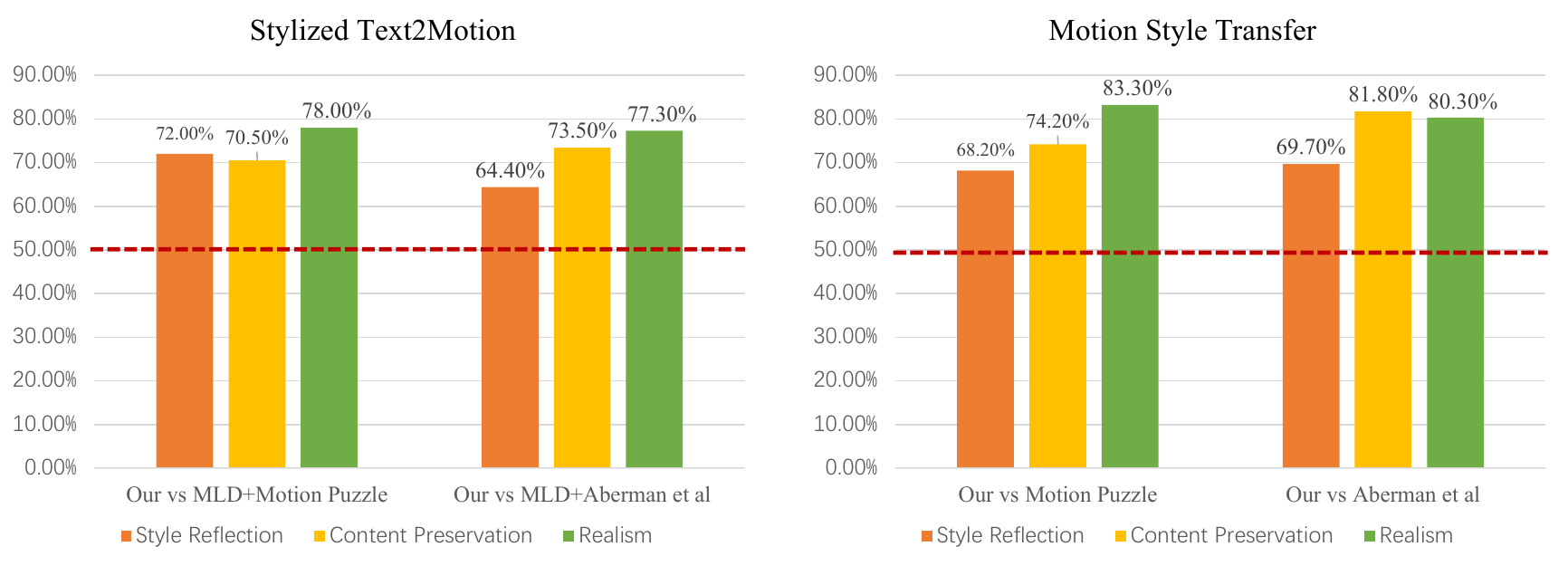}
       \caption{
            \textbf{User Study} on two stylized motion generation tasks. 
           }
       \label{fig:user_study}
\end{figure}

\noindent \textbf{User Study.}
Due to the highly subjective nature of stylized motion, we conduct User studies using pairwise comparisons to further evaluate our proposed method in the tasks of stylized motion generation and motion style transfer.
We recruited 22 human subjects to participate in the study.
In each test, participants are presented with two 4-second video clips synthesized by our method and one comparison method. 
They are then required to select their preferred clip while considering \textit{Realism}, \textit{Style Reflection}, and \textit{Content Preservation} dimensions, respectively.
As shown in Fig.~\ref{fig:user_study}, our method receives more user appreciation compared to two baselines across three dimensions in two tasks.
Further user study details are provided in supplementary.

\subsection{Ablation studies}
To validate the effectiveness of our framework's design choices, we have conducted several ablation studies: the first assesses the impact of each loss function term, while the second evaluates the influence of the style adaptor and style guidance during sampling.%

\noindent \textbf{Loss Components.}%
Firstly, we exclude the \(cycle\text{-}prior\) term in the loss function, denoted as $w/o \; L_{cyc}$.
Comparing the results in the \(1^{\text{st}}\) and \(2^{\text{nd}}\) rows in Table~\ref{tab:ablation}, we observe that our full model outperforms in all content preservation and style reflection metrics.
The motion generated by our approach can still perform the content adhering to the text description but performs worse in accurately reflecting the motion style, as reflected by the arms not being fully extended horizontally.

Since the cycle prior-preservation term is built upon the prior-preservation term, it is meaningless to exclude $L_{pr}$ while retaining $L_{cyc}$.
Therefore, we further exclude both the \(prior\text{-}preservation\) and \(cycle\text{-}prior\) term, denoted as \(w/o \, L_{pr}+L_{cyc}\) in the \(3^{\text{rd}}\) row.
By comparing the results in the \(2^{\text{nd}}\) and \(3^{\text{rd}}\) rows of Table~\ref{tab:ablation}, we notice that while the number of SRAs is higher in the third row, other metrics show a significant decline.
Specifically, in terms of the FID metric, performance deteriorates by more than \textbf{229\%}.
Indeed, without \(L_{pr}\), the model tends to lose the ability to translate content text into corresponding motion, a phenomenon named 'content-forgetting' as described in Sec.~\ref{section:loss}.
Fig.~\ref{fig:ablation}(a) validates our observation, showing that the content in the generated motion significantly deviates from the text descriptions and closely resembles the style motion sequence.

\vspace{1em}
\noindent \textbf{Style Adaptor and Style Guidance.}
Initially, we compare our model to a variant without the classifier-based %
guidance, \textit{w/o classifier-based}, to demonstrate its effectiveness. 
The \(5^{\text{th}}\) row of Table~\ref{tab:ablation} presents the results.
Consistent with the findings in~\cite{jang2022motion}, a reasonable trade-off between content preservation and style reflection is observed.
Although classifier-based style guidance may slightly affect the content preservation metrics, it significantly boosts the model's performance in the SRA metric, yielding an impressive \textbf{208\%} improvement.%
Fig.~\ref{fig:ablation}(d) demonstrates that, without classifier-based style guidance, the generated motions can reflect the motion style, yet they still fall short of fully achieving the target style. Classifier-based style guidance can effectively bridge this gap.

Subsequently, as shown in Table~\ref{tab:ablation}, we evaluate a variant without the style adaptor, denoted as \textit{w/o adaptor} (the last row).
In cases where the SRA values are close, the style adaptor improved the FID metric by about 80.46\%.
Fig.~\ref{fig:ablation}(c) shows that the generated motion can greatly perform the 'walk' action while successfully reflecting the style, but fails to perform the 'sit' action.
This indicates that the effectiveness of classifier-based style guidance diminishes when the content text deviates from locomotion-related movements. Relying solely on it may even adversely affect action performance.

\section{Conclusion}
In this work, we introduce the Stylized Motion Diffusion Model, a novel approach that leverages a pre-trained motion diffusion model to facilitate stylized motion generation driven by content text. 
By integrating a style adaptor and style classifier guidance, our method is capable of producing realistic human motions that accurately reflect both the content text descriptions and the desired motion style from motion sequences.
Through detailed ablation studies, we have demonstrated the effectiveness of each component in our framework.

\newpage
\appendix
\section{Appendix}

\mypar{Video.}
We provide a supplemental video, which we encourage the reviewer to watch since motion is critical in our results, and this is hard to convey in a static document. 

\mypar{Code and Model.}
The code, trained model, and re-targeted 100STYLE datasets will be made publicly available upon acceptance.

\subsection{Pseudo Code}
\newcommand{\commentt}[1]{\textcolor{black}{#1}}
\begin{minipage}{1.0\linewidth} 

\begin{algorithm}[H]
\caption{\textbf{\shortname}'s inference}\label{alg:inference}
\begin{algorithmic}[1]
\Require A motion diffusion model $M$ with parameters $\theta_M$, a style adaptor model $A$ with parameters $\theta_A$, style motion sequence $\vs$ (if any), content texts $\vc$ (if any).
\State $\vz_T \sim \mathcal{N}(\mathbf{0}, \mathbf{I})$ \commentt{\small \# Sample from pure Gaussian distribution}
\ForAll{$t$ from $T$ to $1$}    
    \State $\{\vr\} \leftarrow A(\vz_t, t, \vc, \vs; \theta_A)$ \commentt{\small \indent \# \textbf{Style Adaptor model}}
    \State $\epsilon_t \leftarrow M(\vx_t, t, \vc, \{\vr\};\theta_M)$ \commentt{\small \indent \# Model diffusion model}
    \ForAll{$k$ from $1$ to $K$} \commentt{\small \indent \# \textbf{Classifier-based style guidance}}
        \State $\epsilon_t = \epsilon_t  + \tau \nabla_{\vz_t}G(\vz_t, t, \mathbf{s}) $
    \EndFor
    \State $\vz_{t-1} \sim \mathcal{S}\left(\vz_t, \epsilon_t,t \right)$ \commentt{\small \# $S(\cdot,\cdot,\cdot)$ represents the DDIM sampling method~\cite{dhariwal2021diffusion}.}
\EndFor

\State $\vx_0$ = $\mathbf{D}(\vz_0)$
\\
\Return $\vx_0$
\end{algorithmic}
\end{algorithm}

\end{minipage}

\subsection{Motion Style Transfer}
This task involves taking a content motion sequence along with a style motion sequence and then generating a stylized motion sequence. We treat motion style transfer as one of our downstream applications and can enable \shortname to support it without additional training.
Firstly, we adopt the deterministic DDIM reverse process~\cite{song2020denoising} to obtain the noised latent code $\vz_T^{Inv}$ for the content motion sequence. 
The reverse process can be represented at step $t$ as:
\begin{equation}
\vz_{t+1} = \sqrt{\frac{\alpha_{t+1}}{\alpha_t}} \left( \vz_t + \left( \sqrt{\frac{1}{\alpha_{t+1}}} - 1 \right) - \left( \sqrt{\frac{1}{\alpha_t}} - 1 \right) \right) \cdot \varepsilon_{\theta}(\vz_t; t, \vc,\emptyset),
\end{equation}
where $\alpha$ represents the noise scale.
$\vz_T^{Inv}$ can be obtained at the last reverse step $T$. 
We substitute $\vz_T$, which is initially from a pure Gaussian distribution, with the DDIM-reversed latent $\vz_T^{Inv}$ in Alg.~\ref{alg:inference} and adhere to the same inference procedure to integrate the style condition into the motion content sequence throughout the denoising steps.
Because there are fewer denoising steps compared to the stylized text2motion process, we slightly increase the weights of each style guidance. Specifically, the number of denoising steps is $30$, $w_s = 6.5$ and $\tau=-0.4$

\subsection{Implementation details}
\mypar{Training details.}
Our framework is implemented in PyTorch and trained on a single NVIDIA A5000 GPU. We use a batch size of $64$, train for $50$ epochs, and use the AdamW optimizer~\cite{loshchilov2017decoupled} with a learning rate of 1e-5. Training takes about $1$ hour on a single A5000 GPU, totaling 3700 iterations.
During training, we optimize the style adaptor while keeping the parameters of MLD frozen. 
Furthermore, to learn both the unconditioned and conditioned models simultaneously during training, we randomly set the content text $\vc = \emptyset$ and mask out the style motion sequence $\vs$ in the time dimension by $10\%$.
The number of diffusion steps is $1K$ during training while $50$ during interfering.
The weight of classifier-free content guidance $w_c$ is set to $7.5$, classifier-free style guidance $w_s$ is set to $1.5$, and classifier-based style guidance $\tau$ is set to $-0.2$.

\mypar{Model details.}
We select MLD~\cite{chen2023executing} as our pre-trained motion diffusion model and use its pre-trained weights to initialize both MLD and our style adaptor. 
The style adaptor is composed of $4$ Transformer Encoder blocks.
The input process, as shown in Fig.~\ref{fig:pipeline}, primarily involves a CLIP model\cite{radford2021learning} to encode the content text $\vc$ into text embeddings, and linear layers to project the timestep $t$ into time embeddings. These text embeddings are then added to the time embeddings and concatenated with the noisy latent $\vz_t$, serving as input to the subsequent Transformer Encoder in the latent diffusion model.
The style encoder, as illustrated in Fig.~\ref{fig:pipeline}, primarily consists of a single Transformer Encoder designed to encode the style motion sequences $\vs$ into style embeddings. These style embeddings are then added to the concatenated embeddings from the input process and subsequently fed into the next Transformer Encoder within the style adaptor.

\mypar{Style Function details.}
We opt to first train a style classifier, which consists of a one-layer Transformer block, on the 100STYLE dataset for 100 epochs, using ground-truth style labels for supervision. Then, we omit the last fully connected layer to serve as our style function.

\mypar{Baseline details.}
Due to the baselines being trained on a small style motion dataset and using different skeletons, their released pre-trained weights cannot be directly utilized.
We leverage the source code from Motion Puzzle~\cite{jang2022motion} and Aberman et al.\cite{aberman2020unpaired} to implement their methods on the combined dataset, HumanML3D + 100STYLE. For a fair comparison, we replace their 4D rotation with our 6-D rotation-based feature\cite{zhou2019continuity}. 
Given the requirement for style-labeled motion data in Aberman et al.\cite{aberman2020unpaired}, we follow the same process from Motion Puzzle\cite{jang2022motion} to allow Aberman et al.'s approach to bypass this constraint.
Because these baselines are trained from scratch, we increased their training iterations to five times more than ours.

\mypar{Dataset details.}
Due to some style labels in the 100STYLE dataset inherently containing content meanings, like 'jump' and 'kick', which may conflict with the content text in the HumanML3D dataset. For example, style motion about 'kick' will conflict with content text 'a person walks forward and then backward.' To fairly compute the SRA metric, we follow \cite{kim2019perceptual} to categorize style labels in the 100STYLE dataset into six groups: character (CHAR), personality (PER), emotion (EMO), action (ACT), objective (OBJ), and motivation (MOT). Notably, the 'ACT' group contains content meaning; we exclude the 'ACT' group style motion when computing the SRA metric for content text from the HumanML3D dataset. 
It is worth noting that we use all categories of style motion during training.
Table.~\ref{tab:100style} is the detailed grouping of style labels in the 100STYLE dataset.

\newcolumntype{P}[1]{>{\centering\arraybackslash}p{#1}}

\begin{table}[h]
\centering
\caption{The detailed grouping of style labels in the 100STYLE dataset.}
\label{tab:100style}
\begin{tabular}{|c|P{10cm}|}
\hline
\textbf{Category} & \textbf{Label} \\ \hline
CHAR & Aeroplane, Cat, Chicken, Dinosaur, Fairy, Monk, Morris, Penguin, Quail, Roadrunner, Robot, Rocket, Star, Superman, Zombie (15) \\ \hline
PER & Balance, Heavyset, Old, Rushed, Stiff (5) \\ \hline
EMO & Angry, Depressed, Elated, Proud (4) \\ \hline
ACT & kimbo, ArmsAboveHead, ArmsBehindBack, ArmsBySide, ArmsFolded, BeatChest, BentForward, BentKnees, BigSteps, BouncyLeft, BouncyRight, CrossOver, FlickLegs, Followed, GracefulArms, HandsBetweenLegs, HandsInPockets, HighKnees, KarateChop, Kick, LeanBack, LeanLeft, LeanRight, LeftHop, LegsApart, LimpLeft, LimpRight, LookUp, Lunge, March, Punch, RaisedLeftArm, RaisedRightArm, RightHop, Skip, SlideFeet, SpinAntiClock, SpinClock, StartStop, Strutting, Sweep, Teapot, Tiptoe, TogetherStep, TwoFootJump, WalkingStickLeft, WalkingStickRight, Waving, WhirlArms, WideLegs, WiggleHips, WildArms, WildLegs (58) \\ \hline
MOT & CrowdAvoidance, InTheDark, LawnMower, OnHeels, OnPhoneLeft, OnPhoneRight, OnToesBentForward, OnToesCrouched, Rushed (9) \\ \hline
OBJ & DragLeftLeg, DragRightLeg, DuckFoot, Flapping, ShieldedLeft, ShieldedRight, Swimming, SwingArmsRound, SwingShoulders (9)\\ \hline
\end{tabular}
\end{table}

\subsection{Inference times}
To evaluate the inference efficiency of our submodules, full model, and baseline methods for stylized text2motion tasks, we report the average Inference Time per Sentence measured in seconds (AITS)\cite{chen2023executing}, in Table~\ref{table:time}. The AITS is calculated by setting the batch size to $1$ and excluding the time cost for model and dataset loading on an NVIDIA A5000 GPU.
\begin{table*}[h]
\centering
\resizebox{1.0\textwidth}{!}{

\begin{tabular}{c|ccc|c|ccc}
\toprule
   \multicolumn{1}{p{1.3cm}}{\centering Sub- \\ Modules} &  \multicolumn{1}{|p{1.3cm}}{\centering MLD} & \multicolumn{1}{p{1cm}}{\centering w/o \\ adaptor } &   \multicolumn{1}{p{2.5cm}}{\centering  w/o \\ classifier-based  } & \multicolumn{1}{|p{1.3cm}}{\centering Methods \\ \textit{Overall}} & \multicolumn{1}{|p{1.3cm}}{\centering Ours} & \multicolumn{1}{p{2.6cm}}{\centering MLD + \\ Motion Puzzle} & \multicolumn{1}{p{2.6cm}}{\centering MLD + \\ Aberman et al.} \\ 
 \midrule
 Time (s) & 0.2139  & 2.5081 & 0.5563  & Time (s) & 3.1133 & 0.2420 & 0.2275 \\ 
\bottomrule
\end{tabular}

}
\vspace{4mm}
\caption{
\textbf{Inference time.}
We report the Average Inference Time per Sentence (AITS) in seconds for baselines and each submodule of ours on stylized text2motion tasks.
}
\label{table:time}
\end{table*}

\subsection{More details on classifier-based style guidance}
In our experiments, we observed a phenomenon similar to that described in Text2Image~\cite{yu2023freedom}: In the early denoising stages, the generated motion gradually transitions from random movement to motion that adheres to the content text. Once the global motion content is shaped, subsequent denoising stages primarily focus on modifying the local details and enhancing the quality of the motion. Introducing classifier-based style guidance at an early stage not only poses challenges in steering the motion toward the desired style but also affects the motion's adherence to the content text.
Therefore, we apply classifier-based style guidance near the last stage, once the rough outline of the global motion content has been established and the focus shifts to modifying local details.
Moreover, we can iterate classifier-based guidance multiple times $K$ to improve the steered accuracy:
\[
K = \begin{cases} 
K_e & \text{if } T_s < t < T, \\
K_l & \text{if } t \leq T_s.
\end{cases}
\]
We use $K_e = 0$, $K_l = 5$, and $T_s = 300$ in our experiments. 

\begin{figure}[h]
    \centering
       \includegraphics[width=0.8 \linewidth]{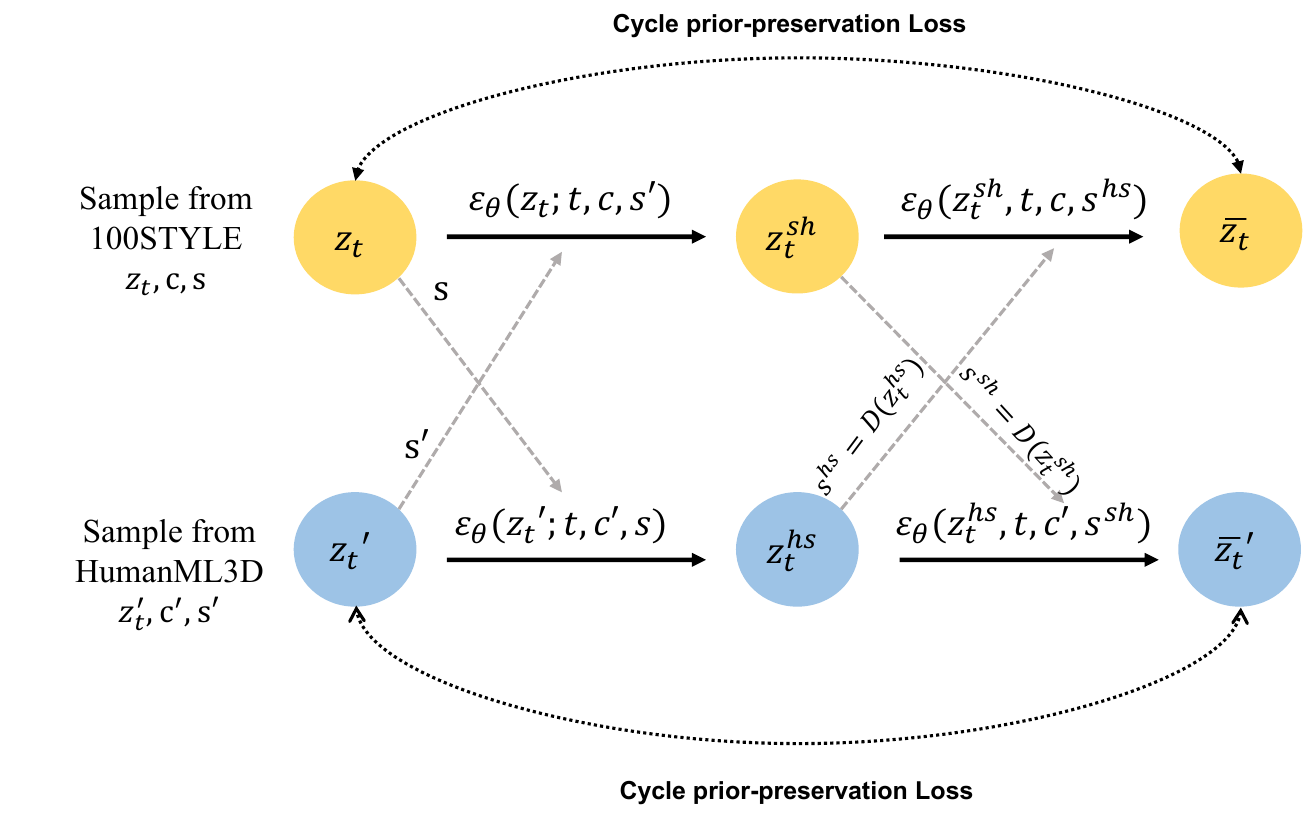}
       \caption{
            \textbf{Visual pipeline} of the cycle prior-preservation loss.
           }
       \label{fig:cycloss}
\end{figure}
\subsection{More details on cycle prior-preservation loss}
We introduce the cycle prior-preservation loss to ensure that generated motion retains content-invariant characteristics from the content text. Fig.~\ref{fig:cycloss} illustrates the cycle prior-preservation loss's visual pipeline.
At timestep $t$, the process begins with sampling content text $\vc$, style motion sequence $\vs$, and noisy motion latent $\vz_t$ from the 100STYLE dataset, alongside their equivalents $\vc^{\prime}$, $\vs^{\prime}$, and $\vz_{t}^{\prime}$ from the HumanML3D dataset. Following this, we facilitate the transfer of content and style conditions between these datasets, yielding $z_t^{sh}$ and $z_t^{hs}$. Decoding $z_t^{hs}$ into the motion space generates the $s^{hs}$ motion sequence. Viewed as a style motion sequence, $s^{hs}$ is combined with the original content text $\vc$ to reconstruct the noisy latent $\Bar{\vz}_t$.
The cycle prior-preservation loss then operates between the original noisy latent $\vz_t$ and the reconstructed noisy latent $\Bar{\vz}_t$.

\subsection{User study details}
To mitigate the potential challenges in participant selection when they are asked to rank or score various methods, we developed an online questionnaire with pairwise A/B tests. We randomly selected 12 sets of stylized motion for the stylized text-to-motion task and 10 sets for the motion style transfer tasks.
We recruited 22 human subjects from various universities, representing a range of academic backgrounds, to participate in our study. At the start of the user study, we introduced the concept of motion stylization, providing examples of both the content text/motion and style motion for reference. With the reference style motion and content text/motion provided, participants were asked to evaluate and choose the better one based on the dimensions of Realism, Style Reflection, and Content Preservation, respectively.
As shown in Fig.~\ref{fig:user_study}, our approach achieves better performance than the baselines on two tasks across three evaluation dimensions.

\subsection{More ablation studies}

\begin{figure}[tp]
    \centering
       \includegraphics[width=1.0\linewidth]{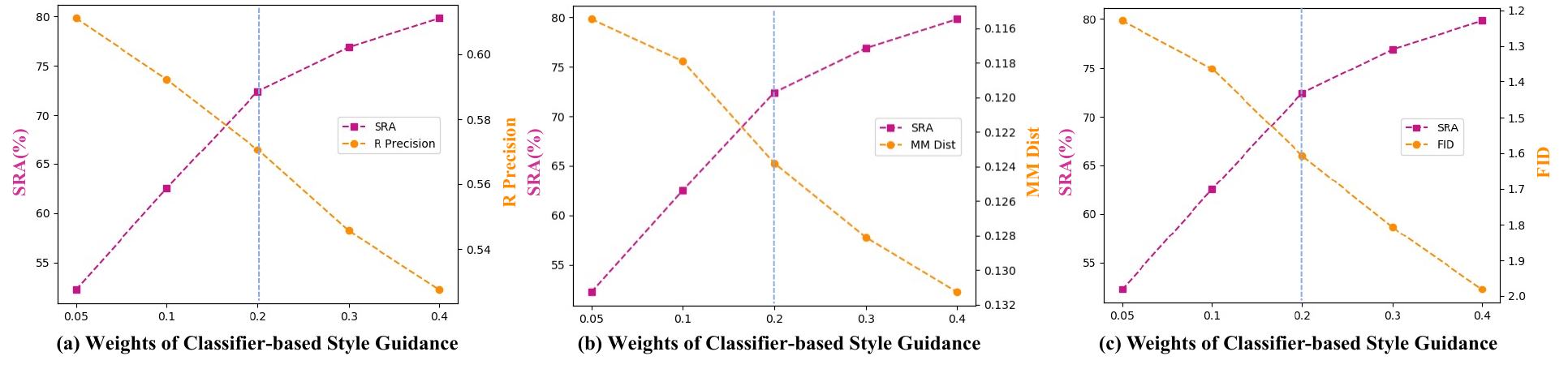}
       \caption{
            \textbf{Varying the weights of the classifier-based style guidance.} 
           }
       \label{fig:trade-off}
\end{figure}

\noindent \textbf{Varying the weight of Classifier-based style guidance.}
\label{sec:cg_weighs}
Due to the flexibility of the style guidance weights, we explore the effects of varying the classifier-based style guidance weight in Fig.~\ref{fig:trade-off}.
We observe that increasing the classifier-based style guidance weight boosts the SRA metric but reduces R Precision, MM Dist, and FID, which means less content preservation but reflecting style more accurately. 
It is observed that when the absolute value of the weight of classifier-based style guidance $\tau$ exceeds $0.2$, the rate of increase for SRA metrics slows down, yet the other metrics continue to deteriorate rapidly. Therefore, we set $\tau = -0.2$ as a trade-off.

\subsubsection{Varying the weights of the classifier-free style guidance.}
\begin{figure}[tp]
    \centering
       \includegraphics[width=1.0\linewidth]{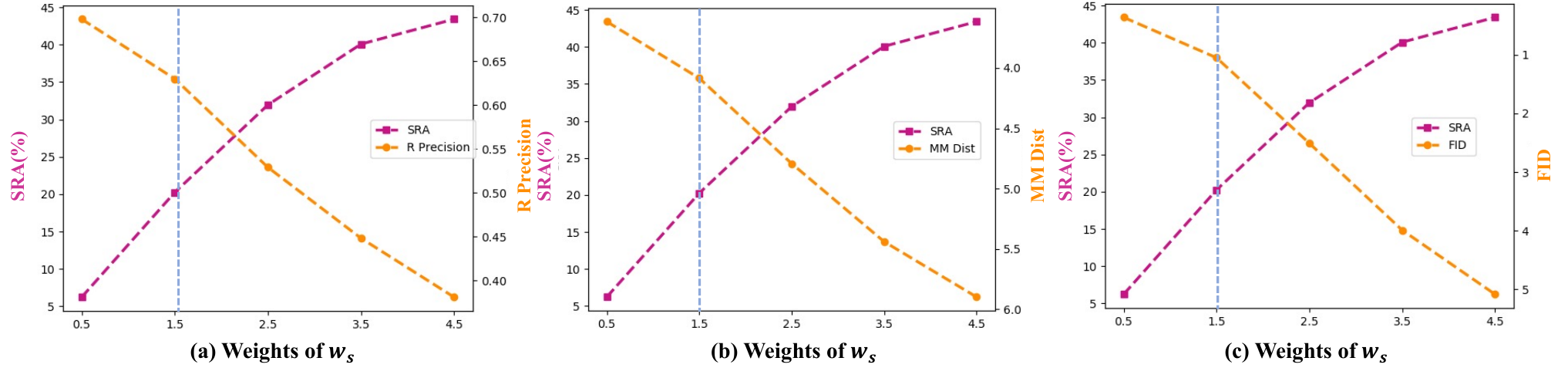}
       \caption{
            \textbf{Varying the weights of the classifier-free style guidance.} 
           }
       \label{fig:weights_cfg}
\end{figure}
Similar to how we can adjust the weights of classifier-based style guidance to balance style reflection and content preservation, as discussed in Sec.~\ref{sec:cg_weighs}, adjusting the weights of classifier-free style guidance also involves a trade-off. Fig. \ref{fig:weights_cfg} illustrates the effects of varying the classifier-free style guidance weights $w_s$, while setting $\tau=0$.
As the weights $w_s$ increase, the SRA gradually increases, while the R-precision and FID metrics deteriorate. It is observed that when $w_s$ exceeds $1.5$, FID, R Precision, and MM Dist decrease more rapidly, whereas SRA continues to increase at the same rate. Therefore, we set $w_s=1.5$ to prevent rapid deterioration in content preservation metrics while ensuring optimal performance in the SRA metric.

\subsubsection{The alternative approach of prior preservation loss.}
In Sec.~\ref{section:loss}, we introduce our prior preservation loss, which involves sampling instances from the HumanML3D dataset as well as from the 100STYLE dataset, and then calculating the loss to prevent 'content-forgetting.' A straightforward alternative approach involves simply combining the 100STYLE and HumanML3D datasets to create a larger dataset, and then only utilizing $L_{std}$ to fine-tune the style adaptor.
Given the larger number of samples in the HumanML3D dataset compared to the 100STYLE dataset, this approach struggles to effectively capture style features from instances in the 100STYLE dataset and maintain learned content in a single optimization step.
We term this alternative method the $\textit{combined dataset}$ approach, utilizing it to train the style adaptor across the same number of training iterations.
Compared to the second and third rows in Table~\ref{tab:ablation_more}, the $\textit{combined dataset}$ approach shows markedly worse performance in content preservation metrics, such as FID and MM Dist values, indicating a failure to preserve content.
These results demonstrate that our simple prior preservation loss can effectively learn style features and simultaneously preserve the learned content with minimal training steps.

\begin{table*}[t]
\centering
\caption{Ablation Studies on HumanML3D Content and 100STYLE Styles.
}
\label{tab:ablation_more}
\resizebox{0.80\textwidth}{!}{
\begin{tabular}{c|cccccccc}
\toprule
Method  & {FID$\downarrow$} & \multicolumn{1}{p{1.8cm}}{\centering Foot skating \\ ratio$\downarrow$} & {MM Dist$\downarrow$}  &\multicolumn{1}{p{1.9cm}}{\centering R-precision$\uparrow$ \\ (Top-3)}& {Diversity$\rightarrow$}& {SRA(\%)$\uparrow$}  \\ 
\midrule
Ours (on all) & 1.609 & 0.124 & 4.477 & 0.571 & 9.235 & 72.418 \\
\midrule
\textit{combined dataset} & 3.892 & 0.332 & 6.152 & 0.379 & 6.833 & 57.573 \\

\bottomrule
\end{tabular}
}
\end{table*}

\begin{figure}[t]
    \centering
       \includegraphics[width=1.00\linewidth]{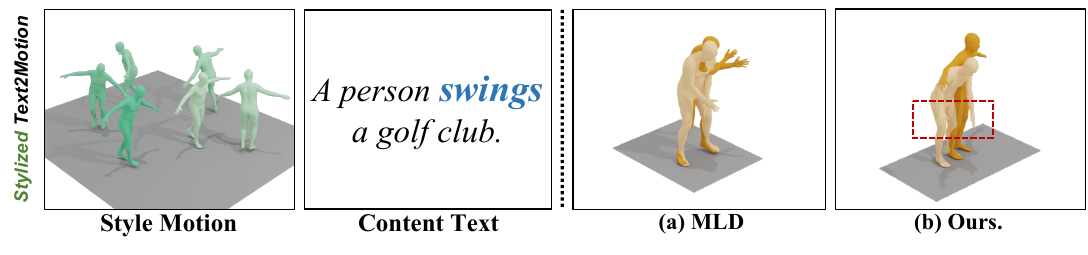}
       \caption{
            A visual example showing conflicts between content text and style motion in a specific body part.
           }
       \label{fig:conflict}
\end{figure}

\begin{figure}[tp]
    \centering
       \includegraphics[width=1.0\linewidth]{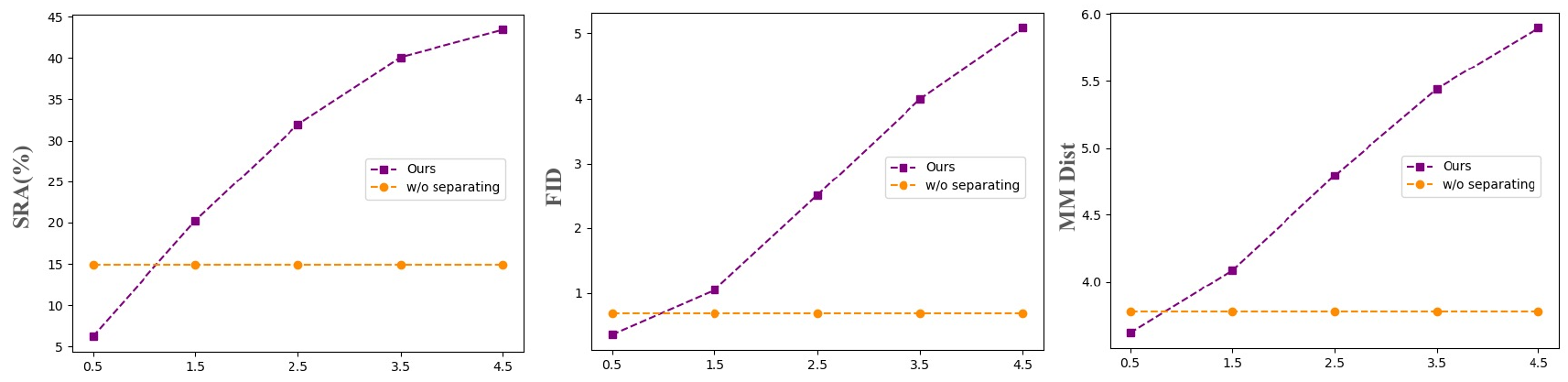}
       \caption{
            \textbf{Comparing our approach with the variant without separating the classifier-free style guidance from content guidance.} 
           }
       \label{fig:compare_wo_cfgs}
\end{figure}

\subsection{Limitation and future plans}
A primary limitation of our approach is its reliance on a pre-trained motion diffusion model, which impacts the realism of the generated motions. 
Consequently, our approach may produce motions with foot skating for certain content texts.
We present these failure cases in the supplementary video.
Incorporating realism guidance~\cite{xie2023omnicontrol} or physical constraints~\cite{yuan2023physdiff} might be a promising direction to improve the realism of the generated motions.

Another limitation is that, due to the classifier-based style guidance potentially requiring iteration, our approach is more time-consuming than MLD by nearly 10 times. A potential direction for improvement involves decreasing the number of denoising steps, inherently reducing the iterations required for classifier-based guidance. Exploring the integration of a one-step model, such as the consistency model~\cite{song2023consistency}, in the motion generation could be a valuable direction.


\begin{thebibliography}{10}
\providecommand{\url}[1]{\texttt{#1}}
\providecommand{\urlprefix}{URL }
\providecommand{\doi}[1]{https://doi.org/#1}

\bibitem{aberman2020unpaired}
Aberman, K., Weng, Y., Lischinski, D., Cohen-Or, D., Chen, B.: Unpaired motion style transfer from video to animation. TOG  (2020)

\bibitem{alexanderson2023listen}
Alexanderson, S., Nagy, R., Beskow, J., Henter, G.E.: Listen, denoise, action! audio-driven motion synthesis with diffusion models. TOG  (2023)

\bibitem{Ao2023GestureDiffuCLIP}
Ao, T., Zhang, Z., Liu, L.: Gesturediffuclip: Gesture diffusion model with clip latents. TOG  (2023)

\bibitem{cen2024generating}
Cen, Z., Pi, H., Peng, S., Shen, Z., Yang, M., Zhu, S., Bao, H., Zhou, X.: Generating human motion in 3d scenes from text descriptions. In: CVPR (2024)

\bibitem{chen2024motionllm}
Chen, L.H., Lu, S., Zeng, A., Zhang, H., Wang, B., Zhang, R., Zhang, L.: Motionllm: Understanding human behaviors from human motions and videos. ArXiv  (2024)

\bibitem{chen2023executing}
Chen, X., Jiang, B., Liu, W., Huang, Z., Fu, B., Chen, T., Yu, G.: Executing your commands via motion diffusion in latent space. In: CVPR (2023)

\bibitem{cohan2024flexible}
Cohan, S., Tevet, G., Reda, D., Peng, X.B., van~de Panne, M.: Flexible motion in-betweening with diffusion models. ArXiv  (2024)

\bibitem{dabral2023mofusion}
Dabral, R., Mughal, M.H., Golyanik, V., Theobalt, C.: Mofusion: A framework for denoising-diffusion-based motion synthesis. In: CVPR (2023)

\bibitem{dai2024motionlcm}
Dai, W., Chen, L.H., Wang, J., Liu, J., Dai, B., Tang, Y.: Motionlcm: Real-time controllable motion generation via latent consistency model. ArXiv  (2024)

\bibitem{dhariwal2021diffusion}
Dhariwal, P., Nichol, A.: Diffusion models beat gans on image synthesis. NeurIPS  (2021)

\bibitem{du2019stylistic}
Du, H., Herrmann, E., Sprenger, J., Fischer, K., Slusallek, P.: Stylistic locomotion modeling and synthesis using variational generative models. In: Proceedings of the 12th ACM SIGGRAPH Conference on Motion, Interaction and Games (2019)

\bibitem{everaert2023diffusion}
Everaert, M.N., Bocchio, M., Arpa, S., S{\"u}sstrunk, S., Achanta, R.: Diffusion in style. In: ICCV (2023)

\bibitem{ghosh2023remos}
Ghosh, A., Dabral, R., Golyanik, V., Theobalt, C., Slusallek, P.: Remos: Reactive 3d motion synthesis for two-person interactions. In: ArXiv (2023)

\bibitem{guo2024momask}
Guo, C., Mu, Y., Javed, M.G., Wang, S., Cheng, L.: Momask: Generative masked modeling of 3d human motions. In: CVPR (2024)

\bibitem{guo2024generative}
Guo, C., Mu, Y., Zuo, X., Dai, P., Yan, Y., Lu, J., Cheng, L.: Generative human motion stylization in latent space. ArXiv  (2024)

\bibitem{Guo_2022_CVPR}
Guo, C., Zou, S., Zuo, X., Wang, S., Ji, W., Li, X., Cheng, L.: Generating diverse and natural 3d human motions from text. In: CVPR (2022)

\bibitem{ho2022classifier}
Ho, J., Salimans, T.: Classifier-free diffusion guidance. ArXiv  (2022)

\bibitem{huang2023diffusion}
Huang, S., Wang, Z., Li, P., Jia, B., Liu, T., Zhu, Y., Liang, W., Zhu, S.C.: Diffusion-based generation, optimization, and planning in 3d scenes. In: CVPR (2023)

\bibitem{jang2022motion}
Jang, D.K., Park, S., Lee, S.H.: Motion puzzle: Arbitrary motion style transfer by body part. TOG  (2022)

\bibitem{jiang2023motiongpt}
Jiang, B., Chen, X., Liu, W., Yu, J., Yu, G., Chen, T.: Motiongpt: Human motion as a foreign language. ArXiv  (2023)

\bibitem{jones2024customizing}
Jones, M., Wang, S.Y., Kumari, N., Bau, D., Zhu, J.Y.: Customizing text-to-image models with a single image pair. ArXiv  (2024)

\bibitem{karunratanakul2023dno}
Karunratanakul, K., Preechakul, K., Aksan, E., Beeler, T., Suwajanakorn, S., Tang, S.: Optimizing diffusion noise can serve as universal motion priors. In: Arxiv (2023)

\bibitem{karunratanakul2023gmd}
Karunratanakul, K., Preechakul, K., Suwajanakorn, S., Tang, S.: Gmd: Controllable human motion synthesis via guided diffusion models. In: ICCV (2023)

\bibitem{kim2019perceptual}
Kim, H.J., Lee, S.H.: Perceptual characteristics by motion style category. In: Eurographics (Short Papers) (2019)

\bibitem{kulkarni2023nifty}
Kulkarni, N., Rempe, D., Genova, K., Kundu, A., Johnson, J., Fouhey, D., Guibas, L.: Nifty: Neural object interaction fields for guided human motion synthesis. ArXiv  (2023)

\bibitem{loshchilov2017decoupled}
Loshchilov, I., Hutter, F.: Decoupled weight decay regularization. ArXiv  (2017)

\bibitem{mason2022local}
Mason, I., Starke, S., Komura, T.: Real-time style modelling of human locomotion via feature-wise transformations and local motion phases. Proceedings of the ACM on Computer Graphics and Interactive Techniques  (2022)

\bibitem{park2021diverse}
Park, S., Jang, D.K., Lee, S.H.: Diverse motion stylization for multiple style domains via spatial-temporal graph-based generative model. Proceedings of the ACM on Computer Graphics and Interactive Techniques  (2021)

\bibitem{peng2023hoidiff}
Peng, X., Xie, Y., Wu, Z., Jampani, V., Sun, D., Jiang, H.: Hoi-diff: Text-driven synthesis of 3d human-object interactions using diffusion models. ArXiv  (2023)

\bibitem{peng2021amp}
Peng, X.B., Ma, Z., Abbeel, P., Levine, S., Kanazawa, A.: Amp: Adversarial motion priors for stylized physics-based character control. TOG  (2021)

\bibitem{petrovich2022temos}
Petrovich, M., Black, M.J., Varol, G.: Temos: Generating diverse human motions from textual descriptions. In: ECCV (2022)

\bibitem{petrovich2024multi}
Petrovich, M., Litany, O., Iqbal, U., Black, M.J., Varol, G., Bin~Peng, X., Rempe, D.: Multi-track timeline control for text-driven 3d human motion generation. In: CVPR (2024)

\bibitem{Pi_2023_ICCV}
Pi, H., Peng, S., Yang, M., Zhou, X., Bao, H.: Hierarchical generation of human-object interactions with diffusion probabilistic models. In: ICCV (2023)

\bibitem{pinyoanuntapong2024bamm}
Pinyoanuntapong, E., Saleem, M.U., Wang, P., Lee, M., Das, S., Chen, C.: Bamm: Bidirectional autoregressive motion model. ArXiv  (2024)

\bibitem{pinyoanuntapong2024mmm}
Pinyoanuntapong, E., Wang, P., Lee, M., Chen, C.: Mmm: Generative masked motion model. In: CVPR (2024)

\bibitem{raab2024monkey}
Raab, S., Gat, I., Sala, N., Tevet, G., Shalev-Arkushin, R., Fried, O., Bermano, A.H., Cohen-Or, D.: Monkey see, monkey do: Harnessing self-attention in motion diffusion for zero-shot motion transfer. ArXiv  (2024)

\bibitem{raab2023single}
Raab, S., Leibovitch, I., Tevet, G., Arar, M., Bermano, A.H., Cohen-Or, D.: Single motion diffusion. ArXiv  (2023)

\bibitem{radford2021learning}
Radford, A., Kim, J.W., Hallacy, C., Ramesh, A., Goh, G., Agarwal, S., Sastry, G., Askell, A., Mishkin, P., Clark, J., et~al.: Learning transferable visual models from natural language supervision. In: ICML (2021)

\bibitem{rempeluo2023tracepace}
Rempe, D., Luo, Z., Peng, X.B., Yuan, Y., Kitani, K., Kreis, K., Fidler, S., Litany, O.: Trace and pace: Controllable pedestrian animation via guided trajectory diffusion. In: CVPR (2023)

\bibitem{ruiz2023dreambooth}
Ruiz, N., Li, Y., Jampani, V., Pritch, Y., Rubinstein, M., Aberman, K.: Dreambooth: Fine tuning text-to-image diffusion models for subject-driven generation. In: CVPR (2023)

\bibitem{shafir2023human}
Shafir, Y., Tevet, G., Kapon, R., Bermano, A.H.: Human motion diffusion as a generative prior. ArXiv  (2023)

\bibitem{shah2023ziplora}
Shah, V., Ruiz, N., Cole, F., Lu, E., Lazebnik, S., Li, Y., Jampani, V.: Ziplora: Any subject in any style by effectively merging loras. ArXiv  (2023)

\bibitem{song2020denoising}
Song, J., Meng, C., Ermon, S.: Denoising diffusion implicit models. ArXiv  (2020)

\bibitem{song2023consistency}
Song, Y., Dhariwal, P., Chen, M., Sutskever, I.: Consistency models. ArXiv  (2023)

\bibitem{tao2022style}
Tao, T., Zhan, X., Chen, Z., van~de Panne, M.: Style-erd: Responsive and coherent online motion style transfer. Arxiv  (2022)

\bibitem{tevet2023human}
Tevet, G., Raab, S., Gordon, B., Shafir, Y., Cohen-or, D., Bermano, A.H.: Human motion diffusion model. In: ICLR (2023)

\bibitem{wan2023tlcontrol}
Wan, W., Dou, Z., Komura, T., Wang, W., Jayaraman, D., Liu, L.: Tlcontrol: Trajectory and language control for human motion synthesis. ArXiv  (2023)

\bibitem{wang2023intercontrol}
Wang, Z., Wang, J., Lin, D., Dai, B.: Intercontrol: Generate human motion interactions by controlling every joint. ArXiv  (2023)

\bibitem{wen2021autoregressive}
Wen, Y.H., Yang, Z., Fu, H., Gao, L., Sun, Y., Liu, Y.J.: Autoregressive stylized motion synthesis with generative flow. In: CVPR (2021)

\bibitem{wu2024motionllm}
Wu, Q., Zhao, Y., Wang, Y., Tai, Y.W., Tang, C.K.: Motionllm: Multimodal motion-language learning with large language models. ArXiv  (2024)

\bibitem{wu2024thor}
Wu, Q., Shi, Y., Huang, X., Yu, J., Xu, L., Wang, J.: Thor: Text to human-object interaction diffusion via relation intervention. ArXiv  (2024)

\bibitem{xia2015realtime}
Xia, S., Wang, C., Chai, J., Hodgins, J.: Realtime style transfer for unlabeled heterogeneous human motion. TOG  (2015)

\bibitem{xie2023omnicontrol}
Xie, Y., Jampani, V., Zhong, L., Sun, D., Jiang, H.: Omnicontrol: Control any joint at any time for human motion generation. In: ICLR (2024)

\bibitem{xu2023adaptnet}
Xu, P., Xie, K., Andrews, S., Kry, P.G., Neff, M., McGuire, M., Karamouzas, I., Zordan, V.: Adaptnet: Policy adaptation for physics-based character control. TOG  (2023)

\bibitem{xu2024cyclenet}
Xu, S., Ma, Z., Huang, Y., Lee, H., Chai, J.: Cyclenet: Rethinking cycle consistency in text-guided diffusion for image manipulation. NeurIPS  (2024)

\bibitem{xu2023interdiff}
Xu, S., Li, Z., Wang, Y.X., Gui, L.Y.: Interdiff: Generating 3d human-object interactions with physics-informed diffusion. In: ICCV (2023)

\bibitem{xu2024interdreamer}
Xu, S., Wang, Z., Wang, Y.X., Gui, L.Y.: Interdreamer: Zero-shot text to 3d dynamic human-object interaction. ArXiv  (2024)

\bibitem{yi2024generating}
Yi, H., Thies, J., Black, M.J., Peng, X.B., Rempe, D.: Generating human interaction motions in scenes with text control. ArXiv  (2024)

\bibitem{yu2023freedom}
Yu, J., Wang, Y., Zhao, C., Ghanem, B., Zhang, J.: Freedom: Training-free energy-guided conditional diffusion model. ArXiv  (2023)

\bibitem{yuan2023physdiff}
Yuan, Y., Song, J., Iqbal, U., Vahdat, A., Kautz, J.: Physdiff: Physics-guided human motion diffusion model. In: ICCV (2023)

\bibitem{zhang2023adding}
Zhang, L., Rao, A., Agrawala, M.: Adding conditional control to text-to-image diffusion models. In: ICCV (2023)

\bibitem{zhang2024motiondiffuse}
Zhang, M., Cai, Z., Pan, L., Hong, F., Guo, X., Yang, L., Liu, Z.: Motiondiffuse: Text-driven human motion generation with diffusion model. PAMI  (2024)

\bibitem{zhang2023motiongpt}
Zhang, Y., Huang, D., Liu, B., Tang, S., Lu, Y., Chen, L., Bai, L., Chu, Q., Yu, N., Ouyang, W.: Motiongpt: Finetuned llms are general-purpose motion generators. ArXiv  (2023)

\bibitem{zhou2023emdm}
Zhou, W., Dou, Z., Cao, Z., Liao, Z., Wang, J., Wang, W., Liu, Y., Komura, T., Wang, W., Liu, L.: Emdm: Efficient motion diffusion model for fast, high-quality motion generation. Arxiv  (2023)

\bibitem{zhou2019continuity}
Zhou, Y., Barnes, C., Lu, J., Yang, J., Li, H.: On the continuity of rotation representations in neural networks. In: CVPR (2019)

\end{thebibliography}
\end{document}